\documentclass[journal]{IEEEtran}
\usepackage{epsfig}
\usepackage{graphicx}
\usepackage{amsmath}
\usepackage{amssymb}
\usepackage{mathtools}
\usepackage{enumitem}
\usepackage{booktabs}

\usepackage{amssymb}

\usepackage{pifont}
\usepackage{url}
\usepackage{varioref}
\usepackage[symbol]{footmisc}

\usepackage[dvipsnames]{xcolor}
\usepackage{utils}

\usepackage{makecell}
\usepackage{multirow}
\usepackage{bbm}
\usepackage{wrapfig}
\usepackage{booktabs} 
\usepackage{siunitx} 
\usepackage{multirow} 
\usepackage{multicol} 
\usepackage{amssymb} 
\usepackage{adjustbox}
\usepackage{algorithm}
\usepackage{algorithmic}
\usepackage{indentfirst} %

\usepackage{amsmath,amsfonts}
\usepackage{algorithmic}
\usepackage{algorithm}
\usepackage{array}
\usepackage[caption=false,font=normalsize,labelfont=sf,textfont=sf]{subfig}
\usepackage{textcomp}
\usepackage{stfloats}
\usepackage{url}
\usepackage{verbatim}
\usepackage{graphicx}
\usepackage{cite}
\usepackage[numbers, sort & compress]{natbib}
\usepackage{tabularx}
\usepackage{booktabs}
\usepackage{color}
\usepackage{amssymb}
\usepackage{bbm}
\usepackage{xspace}
\usepackage{multirow}
\usepackage[colorlinks,
            linkcolor=green,       %
            anchorcolor=green,  %
            citecolor=green,        %
            ]{hyperref}
\usepackage{cleveref}
\usepackage{hhline}
\usepackage{vcell}

\ifCLASSINFOpdf
\else
\fi

\begin{document}
%
\title{REPAIR: Rank Correlation and Noisy Pair Half-replacing with Memory for Noisy Correspondence}
%
%
%

\author{Ruochen~Zheng,
        Jiahao~Hong,
        Changxin~Gao,
        Nong~Sang
\thanks{Manuscript received XXX; revised XXX. This work was supported by the National Natural Science Foundation of China No.62176097, Hubei Provincial Natural Science Foundation of China No.2022CFA055.
(\textit{Corresponding Author: Changxin Gao})}
\thanks{R. Zheng, J. Hong, C. Gao and N. Sang are with National Key Laboratory of Multispectral Information Intelligent Processing Technology, School of Artificial Intelligence and Automation, Huazhong University of Science and Technology. E-mail: \{ruocz, hongjiahao, cgao, nsang\}@hust.edu.cn.}
}

%
%

\markboth{SUBMITTED TO IEEE TRANSACTIONS ON IMAGE PROCESSING}%
{Zheng \MakeLowercase{\textit{et al.}}: REPAIR: Rank Correlation and Noisy Pair Half-replacing with Memory for Noisy Correspondence}
%



\maketitle

\begin{abstract}
The presence of noise in acquired data invariably leads to performance degradation in cross-modal matching.
Unfortunately, obtaining precise annotations in the multimodal field is expensive, which has prompted some methods to tackle the mismatched data pair issue in cross-modal matching contexts, termed as noisy correspondence.
However, most of these existing noisy correspondence methods exhibit the following limitations:
a) the problem of self-reinforcing error accumulation, and 
b) improper handling of noisy data pair.
To tackle the two problems, we propose a generalized framework termed as Rank corrElation and noisy Pair hAlf-replacing wIth memoRy (REPAIR), which benefits from maintaining a memory bank for features of matched pairs.
Specifically, we calculate the distances between the features in the memory bank and those of the target pair for each respective modality, and use the rank correlation of these two sets of distances to estimate the soft correspondence label of the target pair.
Estimating soft correspondence based on memory bank features rather than using a similarity network can avoid the accumulation of errors due to incorrect network identifications.
For pairs that are completely mismatched, REPAIR searches the memory bank for the most matching feature to replace one feature of one modality, instead of using the original pair directly or merely discarding the mismatched pair.
We conduct experiments on three cross-modal datasets, \textit{i.e.}, Flickr30K, MS-COCO, and CC152K, proving the effectiveness and robustness of our REPAIR on synthetic and real-world noise.
\end{abstract}


\section{Introduction}
\label{sec:intro}
Cross-modal matching, which refers to establishing connections among different modalities, has attracted much attention with the growing presence of multimedia data in daily life.
The paradigm of cross-modal matching is to project different modalities onto a unified space where querying and comparison can be performed.

\begin{figure}[t]
    \centering
    \includegraphics[width=0.47\textwidth]{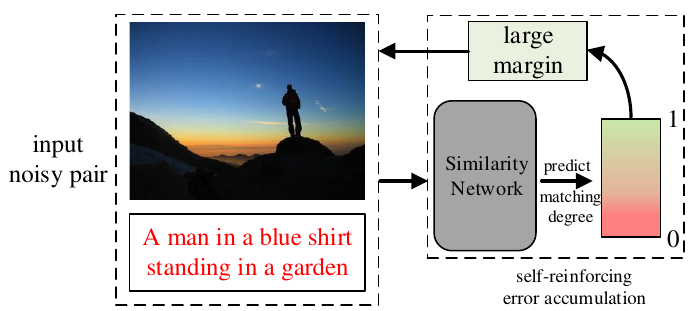}  \\
    (a) \\ 
    \vspace{0.5cm} 
    \includegraphics[width=0.48\textwidth]{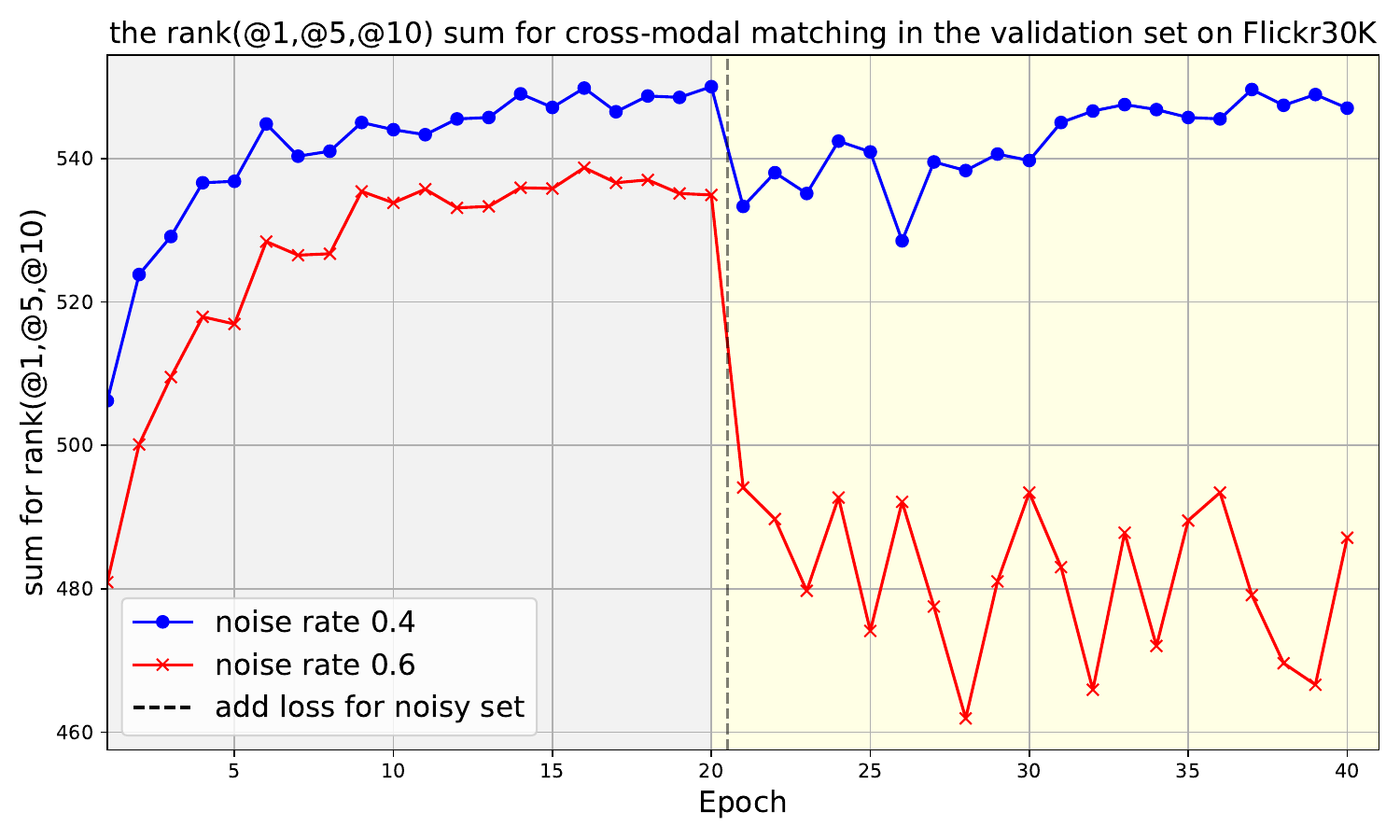} \\
    (b)
    \caption{The issues with the existing methods. (a) The problem of self-reinforcing error accumulation. When a mismatched data pair is mistakenly predicted as high soft correspondence, which is then transformed into a stricter soft margin, it leads to the accumulation of errors. (b) Performance trend of the validation set during the NCR\cite{huang2021learning} training process. Notably, after the 20-th epoch, a loss associated with the noisy set is introduced, leading to a rapid decline in performance, especially in the setting of 0.6 noise rate.}
    \label{fig:two_images}
\end{figure}

Many existing cross-modal methods\cite{faghri2017vse++,radford2021learning,kim2021vilt,diao2021similarity,zhang2020context,chen2020uniter,gabeur2020multi} have succeeded with deep learning.
Most of these methods typically share a fundamental assumption: the data gathered for deep learning training is properly aligned.
However, accurately annotating multimodal data demands significant human and financial resources, given the inherent semantic differences between modalities.
An alternative option is to use unverified multimodal data directly. For instance, in image-text matching, one can inexpensively collect numerous unreliable image-text pairs from the internet\cite{radford2021learning,sharma2018conceptual}, many of which are weakly matched or mismatched pairs.
This situation, where paired data contains alignment errors, is called \emph{noisy correspondence}\cite{huang2021learning}. Under data conditions of noisy correspondence, the performance of current cross-modal matching methods\cite{lee2018stacked,diao2021similarity,chen2021learning} will witness a substantial drop.

Some methods\cite{huang2021learning,feng2023learning,han2023noisy,qin2022deep,qin2023cross} have already noticed this issue and proposed improvements for noisy correspondence.
The core of these methods can be summarized as estimating the matching degree (also termed soft correspondence label) of each data pair, which will recast as soft margins to train the cross-modal matching model.
However, these methods have the following limitations: (a) the computation of soft correspondence label is not directly based on inter-modal features but is derived from a similarity network, which suggests the risk of cyclical self-reinforcing error accumulation, (b) for low correspondence pairs (noisy subset), current methods simply reduce its soft margin constraint or directly discard it.
As shown in Figure~\ref{fig:two_images} (a), methods\cite{huang2021learning,feng2023learning} evaluate the soft correspondence based on the network's prediction results. If the similarity network mistakenly identifies a mismatched pair as a closely matched pair, methods\cite{huang2021learning} consistently assigns a large soft margin to it, which suggests the noisy pair is likely to be recognized as a matched pair in the following training epoch, leading to cumulative errors.
And the negative impact of low soft correspondence pairs on performance is illustrated in Figure~\ref{fig:two_images} (b).
There's a marked decrease in performance when the loss for low correspondence pairs is incorporated, even though these pairs are allocated lower soft margins by current methods.
Although Han~\etal\cite{han2023noisy} propose a strategy by discarding these pairs, we argue that such a strategy is suboptimal as it results in losing information.

To address the aforementioned issues, we propose a general framework for noisy correspondence, named \textbf{R}ank corr\textbf{E}lation and \textbf{P}air h\textbf{A}lf-replacing w\textbf{I}th memo\textbf{R}y (REPAIR), which benefits from employing the memory bank.
Following the previous methods\cite{huang2021learning,qin2022deep,yang2023bicro}, we divide the dataset into noisy and clean subsets, depending on the values of loss.
REPAIR maintains a memory bank using features extracted from data pairs in the clean subset, as the majority of pairs within it are matched.
To avoid the error accumulation of similarity network, REPAIR utilizes the memory bank to evaluate the soft correspondence label of the target data pairs. 
The core idea of this evaluation can be summarized as: \emph{similar relationships in one modality should mirror those in its connected modality}.
A straightforward example can clarify this idea: if a target pair $(I_i,  T_i)$ is matching, then for another two matched pairs $(I_j, T_j)$ and $(I_m, T_m)$, if the distance between $(I_i, I_j)$ is less than that between $(I_i, I_m)$, then the distance between $(T_i, T_j)$ should be less than $(T_i, T_m)$, otherwise, $(I_i, T_i)$ might not be matched.
Taking image-text matching as an instance, the target image computes its distance with the image features in the memory bank to get a group of image distances, and the target text modality operates similarly.
Thus, if the target image-text aligns, two groups of distances across image modality and text modality should consistently align in the rank.
We use \emph{Spearman's rank correlation} to evaluate the correlation between the two groups of distances and then convert it into the soft correspondence label for the target image-text pair.

As for the noisy subset, REPAIR adopts a \emph{noisy pair half-replacing} (NPR) strategy to utilize mismatched data pairs more efficiently. 
Specifically, if the pair is highly likely mismatched, we replace one of the modalities to obtain a new pair assisted by the memory bank.
For instance, if we replace the image while preserving the text, we will search the memory bank for a set of candidate texts closely related to the target text, and from the image features corresponding to the candidate texts, select the image feature most aligned with the original text.
Finally, the newly obtained pairs will participate in training to maximize data utilization.
The highlights of our contributions can be summarized as:
\begin{itemize}
\item In response to the noisy correspondence challenge, we propose a general framework named RAPAIR, which employs a memory bank to maintain a collection of clean pairs, and adopts the \emph{rank correlation} (RC) strategy based on the memory bank to evaluate the correspondence label of target pairs.
\item For entirely mismatched data pairs, REPAIR proposes the \emph{noisy pair half-replacing} (NPR) strategy, which selects appropriate feature from the memory bank to substitute the feature of one modality in the target pair, creating a temporary feature pair with higher matching degree for training.
NPR prevents performance degradation due to the use of wholly misaligned data and avoids the wastage of simply discarding information.
\item The experiments on three challenging cross-modal datasets, \textit{i.e.}, Flickr30K, MS-COCO, and CC152K, prove the effectiveness of REPAIR.
\end{itemize}

\section{Related work}
\subsection{Cross-model Matching}
Cross-modal matching aims to project different modalities onto a unified space where querying and comparison can be performed.
Most existing methods, based on their use of detailed similarity comparisons, can be categorized into two classes: coarse-grained alignment\cite{faghri2017vse++,sarafianos2019adversarial,wang2018learning} and fine-grained alignment\cite{lee2018stacked,chen2020uniter,kim2021vilt,li2019visual,diao2021similarity}.
Coarse-grained alignment methods establish metric relationships between global features of different modalities.
For example, VSE++\cite{faghri2017vse++} employs hard negative mining techniques to enhance the discriminative ability of the global features extracted by the model.
TIMAM\cite{sarafianos2019adversarial} utilizes adversarial learning and pre-trained language models to aid in establishing modality-invariant features.
In contrast, fine-grained alignment techniques, when measuring the similarity of different modalities, take the relationship of local information across those modalities into account by designing the similarity network.
For instance, in the case of image-text matching, the semantic relationships between different objects in the image and words in the sentence are captured by the similarity function.
Some methods\cite{lee2018stacked,chen2020uniter,kim2021vilt} employ attention mechanisms to explore the connections between regions of image and words in the text.
Besides, Graph Convolutional Network (GCN)\cite{kipf2016semi} is also a typical approach for such fine-grained similarity measurements.
Li \etal propose a reasoning model\cite{li2019visual} that uses GCN and memory mechanisms to generate visual representations by capturing key objects in a scene.
Furthermore, Diao \etal\cite{diao2021similarity} introduce the Similarity Graph Reasoning and Attention Filtration (SGRAF) network for image-text matching, which learns alignments using a graph neural network and filters them for significance.

Most of these approaches assume that data pairs are correctly matched. However, in real-world scenarios, the collected data is often mismatched or weakly matched\cite{sharma2018conceptual}, making it challenging for these cross-modal methods to be directly applied in real-world scenarios.
\subsection{Memory-based Learning}
In recent years, the memory bank has already seen widespread application in deep learning\cite{he2020momentum,chen2020improved} and computer vision\cite{li2021memorize,dou2022gaitmpl,lin2023prototypical,li2021align,yang2023membridge,dai2021dual,wang2022delving,yang2023joint,li2022cluster,yin2023real}.
The role of the memory bank in these methods is to store features, which allows for efficient comparison of features between batches.
For instance, in self-supervised learning, memory banks are utilized to expand the negatives, enabling the acquisition of a larger negative sample space with lower memory consumption, thereby enhancing the model's performance.
In some classification or feature matching tasks, \textit{e.g}, reid\cite{zheng2019camera,zhao2023content}, the memory bank\cite{li2022cluster,yin2023real} is commonly used to store features of certain classes as prototypes, which are then utilized to aid the training of the current iteration.
As a core component, the memory bank also plays a vital role in our REPAIR.
On one hand, the memory bank stores clean pair features and utilizes them in the RC module to evaluate the matching degree of each data pair. On the other hand, within the NPR module, it employs these features to replace the completely mismatched pairs and form new feature pairs.

\subsection{Noisy Label Learning}
Faced with a vast amount of data with unreliable labels in real-world scenarios, the process of accurately relabeling such noisy data and then training models on clean data comes at a significant expense.
Therefore, numerous methods\cite{huang2023twin,li2020dividemix,bucarelli2023leveraging,tu2023learning,han2019deep,mao2022noise,huang2020self,reed2014training,arazo2019unsupervised,song2019selfie,mao2023noise} have focused on training deep networks under noisy environments, commonly known as \emph{noisy label learning}.
Here, we mainly discuss the methods related to label adjustment methods that closely align with our approach.
Arazo \etal employ a beta mixture model to determine the probability of a sample being noisy or clean\cite{arazo2019unsupervised}, and dynamically adjust the label by combining this probability with the prediction of the network.
Divide Mix\cite{li2020dividemix} adopts a co-teaching strategy\cite{han2018co} where two networks mutually select samples for each other, and for the noisy samples, the pseudo-label is determined by averaging the outputs of the two networks.
Alternatively, certain techniques\cite{tanaka2018joint,huang2020self} utilize historical output from the training process to generate soft labels for noisy samples.
Some approaches also begin by extracting features from noisy samples and then generate pseudo-labels via K-Nearest Neighbor\cite{li2022selective,ortego2021multi}.
However, most of these methods are based on classification problems and cannot apply to the noise in multimodal field.

\subsection{Noisy Correspondence Learning}
Noise often emerges in multimedia data due to mismatches among different modalities, such as in image-text matching\cite{huang2021learning, qin2023noisy}, infrared re-identification\cite{yang2022learning}, visual-audio matching\cite{han2023noisy}, text-to-video retrieval\cite{amrani2021noise} and multi-view learning\cite{yang2021partially, yang2022robust}.
Noisy correspondence was first introduced by Huang \etal\cite{huang2021learning}, referring to alignment errors present in multimodal data.
To solve this problem, Noisy Correspondence Rectifier (NCR)\cite{huang2021learning} employs network predictions to modify the soft labels, reflecting the matching degree of the target pair.
Subsequently, based on the matching degree of the positive pair, the margin for the triplet loss associated with the pair is dynamically adjusted.
Expanding on this concept, Yang \etal propose Bi-directional Cross-modal Similarity Consistency (BiCro) to estimate the soft correspondence guided by the idea \emph{similar images should have texts and vice versa}.
Han \etal\cite{han2023noisy} employ an additional clean dataset and meta network to measure the correspondence of data pair.
Conversely, Feng \etal categorize the dataset into clean, noisy, and hard subsets\cite{feng2023learning}, designing a more refined soft margin for different subsets.
Recently, Qin \etal propose a strategy\cite{qin2023cross} by aggregating historical predictions to provide stable soft correspondence estimation.
Xu~\etal\cite{xu2023negative} employ clip\cite{radford2021learning} and memory bank for the evaluation of each data pair's performance variation following training, and then use distinct weights to train each pair individually.
Besides the previously mentioned methods, DECL\cite{qin2022deep}, which leverages uncertainty, and RCL\cite{hu2023cross}, adopting the negative learning strategy, have also achieved promising performance in noisy correspondence, while Amrani \etal employed a density-based\cite{amrani2021noise} method to evaluate the matching degree of noisy pairs.

The study most closely related to our REPAIR is BiCro\cite{yang2023bicro}, which generates soft correspondence labels for target pairs by calculating the distance ratio between the nearest clean pair within a batch across two modalities. 
In contrast, our method utilizes a memory bank and focuses on rank correlation, rather than depending on distance ratios.
Employing rank correlation presents the following distinctions: a). it emphasizes the relative relationships of distances rather than the numerical values of the distances themselves, b). it focuses on the correlation of the whole memory bank instead of specific individual pairs.
Furthermore, our REPAIR proposes a novel NPR strategy for mismatched pairs, enabling more efficient use of these data to enhance performance.

\begin{figure*}[ht]
\scriptsize
\centering
\renewcommand{\tabcolsep}{1pt} 
\renewcommand{\arraystretch}{1} 
\begin{center}
\begin{tabular}{cc}
  \includegraphics[width=0.24\linewidth]{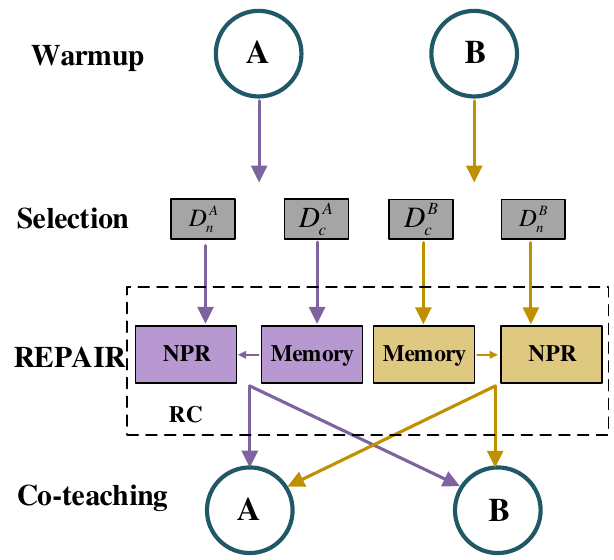} &
  \includegraphics[width=0.72\linewidth]{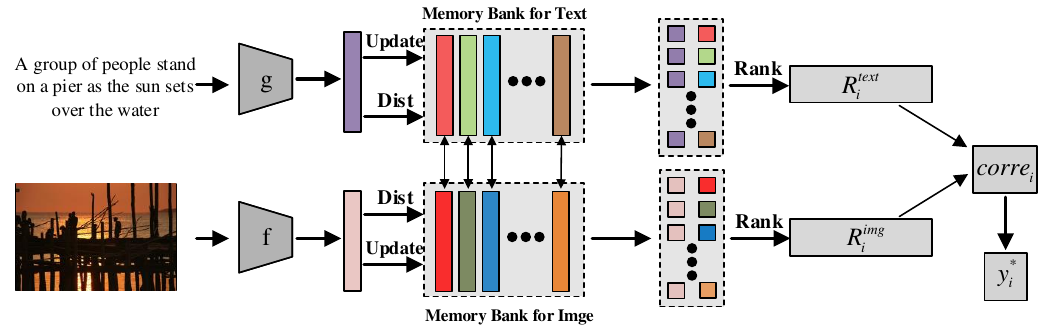}\\   (a) & (b)\\
\end{tabular}
\end{center}
\vspace{-2mm}
\caption{(a) The training pipeline for REPAIR. \emph{Rank correlation} and \emph{noisy pair half-replacing} are abbreviated as RC and NPR, respectively.(b) Illustration of memory bank update for clean set and the rank correlation to obtain the soft correspondence label.}  
\label{fig:pipeline}
\end{figure*}

\section{Our method}
\subsection{Problem Formulation}
Without compromising generality, we use image-text matching to illustrate the cross-modal matching task and the issue of noisy correspondence.
Given a training dataset $\Tilde{D} = \{(I_{i},T_{i}), \Tilde{y_{i}}\}_{i=1}^{N}$, where $(I_{i},T_{i})$ is the $i$-th image-text pair in the dataset and $N$ is the number of pairs. 
And $\Tilde{y_{i}} \in \{0,1\}$ is the binary label indicating whether the pair $(I_{i},T_{i})$ is correctly matched.
However, all the collected data are assumed to be positive pairs $(\Tilde{y_{i}} = 1)$, whereas some of them are misaligned (ground truth $y_{i}=0$).
The challenge faced by cross-modal matching under such data conditions is termed \emph{noisy correspondence}\cite{huang2021learning}.

Traditionally, the goal of cross-modal matching task is to maximize the similarity of positive pairs and minimize that of negative pairs.
Without loss of generality, we denote $f$ and $g$ as the encoders for extracting features from image and text modalities, respectively, and employ a similarity function 
$S(f(I),g(T))$ to estimate their similarity.
For convenience, we will use $S(I,T)$ to denote this similarity function in the subsequent sections of this paper.

In the previous studies\cite{huang2021learning, yang2023bicro, feng2023learning,han2023noisy}, a common approach was to enhance the triplet loss function by estimating the soft correspondence label $y_{i}^{*} \in [0,1]$ and adjusting the soft margin. The enhanced triplet loss can be depicted as:
\begin{equation}
\begin{aligned}
L_{tri}(I_i, T_i) &= \left[ \hat{\alpha}_i - S(I_i, T_i) + S(I_i, \hat{T}_h) \right]_{+} \\
& + \left[ \hat{\alpha}_i - S(I_i, T_i) + S(\hat{I}_h, T_i) \right]_{+}
\label{eq1}
\end{aligned}
\end{equation}
where $[x]_{+} = max(x,0)$, and $\hat{\alpha}_i$ is the soft margin according to $y_{i}^{*}$, $\hat{T}_h  = {argmax}_{T_{j} \neq T_{i}} S(I_{i},T_{j})$ and $\hat{I}_h = {argmax}_{I_{j} \neq I_{i}} S(I_{j},T_{i})$ are hard negatives in a mini-batch.
In NCR\cite{huang2021learning}, an exponential form of soft labels is proposed, allowing more similar positive pair to have a larger margin, which suggests stricter constraint:
\begin{equation}
\begin{aligned}
\hat{\alpha}_{i} = \frac{m^{y^{*}_i} - 1}{m-1} \alpha 
\label{eq2}
\end{aligned}
\end{equation}
where $m$ is the hyperparameter and $\alpha$ is the fixed margin.
Therefore, the research for noisy correspondence shifts to how to estimate the soft correspondence labels $y^*_i$ of the target positive pair\cite{yang2023bicro}.
In our REPAIR, we will continue to focus on the estimation of soft correspondence label to address the noisy correspondence challenge.

Some modules are also validated for their effectiveness in previous methods\cite{huang2021learning,yang2023bicro,feng2023learning}. To obtain better performance, we continue to adopt them in this paper: \emph{warm up}, \emph{co-teaching}\cite{han2018co}, and \emph{sample selection}.
\emph{Warm up} refers to not directly adopting hard negative mining, but instead, initially calculating the triplet loss using an average across all negative pairs. 
\emph{Sample selection} entails dividing the dataset into a clean and noisy subset before each training epoch, based on either loss values. NCR\cite{huang2021learning} implements a two-component Gaussian mixture model for this purpose, whereas method BiCro\cite{yang2023bicro} favors a Beta-Mixture-Model.
\emph{Co-teaching} denotes employing two networks mutually selecting high-confidence pairs to train each other, which can prevent the accumulation of errors brought about by self-selection.

Following the settings in \cite{huang2021learning,feng2023learning}, we simultaneously train two networks $A = \{f^A,g^A,S^A\}$ and $B = \{f^B,g^B,S^B\}$, mutually selecting clean pairs based on the posterior probabilities $w_{i}$ predicted by the Gaussian mixture model, which is based on the per-sample loss of each sample pair within a batch.
Specifically, given a data pair $(I_i, T_i)$, the loss for it in a batch can be defined as:
\begin{equation}
\begin{aligned}
L_{w}(I_i, T_i) &= \sum_{\Tilde{T}}{} \left[ \alpha - S(I_i, T_i) + S(I_i, \Tilde{T}) \right]_{+} \\
& + \sum_{\Tilde{I}}{}\left[ \alpha - S(I_i, T_i) + S(\Tilde{I}, T_i)\right]_{+}\\
\label{eq11}
\end{aligned}
\end{equation}
where $\alpha$ is the fixed margin, set at 0.2.
And $\Tilde{T}$ and $\Tilde{I}$ present the negatives in a batch.
For each data pair, by employing Eq~\ref{eq11}, we can derive a set of loss, denoted as $\{l_i\}_{i=1}^{N} = \{L_w(I_i, T_i)\}_{i=1}^{N}$.

Next, we employ a two-component Gaussian Mixture Model (GMM)\cite{li2020dividemix,permuter2006study} to fit the per-sample loss.
The probability density function (pdf) for a two-component mixture model applied to the loss is can be defined as follows:
\begin{equation}
\begin{aligned}
p(l) = \sum_{t=1}^{2} {\beta}_t p(l|t)
\label{eq12}
\end{aligned}
\end{equation}
where $p(l|t)$ represents the the probability density of the $t$-th component, and ${\beta}_t$ is the corresponding mixture coefficient.
To optimize the GMM, the Expectation-Maximization algorithm will be employed.
Finally, the probability of $(I_i, T_i)$ being clean can be expressed as:
\begin{equation}
\begin{aligned}
w_i = p(k|l_i) = \frac{p(k)p(l_i|k)}{p(l_i)}
\label{eq13}
\end{aligned}
\end{equation}
where $k$ represents the Gaussian component with the smallest mean.
Since we employ per-sample loss to evaluate the clean probability of each sample, the Gaussian component with a lower mean is considered clean, whereas the other is deemed noisy.
By setting a threshold $p$, we can partition dataset $\Tilde{D}$ into clean subset $D_{c}^{k} = \{(I_{i},T_{i}) | w_i^k > p\}$ and noisy subset $D_{n}^{k} = \{(I_{i},T_{i}) | w_i^k <= p\}$, where $k \in \{A,B\}$.
Such a partition is executed before the beginning of each training epoch, and the training pipeline can be found in Figure~\ref{fig:pipeline} (a).

\subsection{Memory Bank Based Rank Correlation}
Compared to previous methods, our REPAIR benefits from maintaining a feature memory bank from clean subset $D_{c}^{k}$.
For the data pairs $(I_{i},T_{i})$ from $D_{c}^{k}$,  after extracting features by $(f^k, g^k)$, these features are pushed into the corresponding network's memory bank $Memory^{k}\{(f_i,g_i)\}_{i=1}^{M}$ (abbreviated as \emph{Memory}), while the earliest stored features are popped out.
$f_i$ and $g_i$ is the abbreviation of $f^{k}(I_i)$ and $g^{k}(T_i)$, and $M$ is the bank size of the memory bank.
In this way, we obtain a dynamically updated feature bank derived from the $D_{c}^{k}$ for both networks $A$ and $B$. In the following sections, we will utilize this memory bank to evaluate the soft correspondence for target pairs, as shown in Figure~\ref{fig:pipeline} (b).

Our core concept can be expressed as: \emph{similar relationships in one modality should mirror those in its connected modality}.
Specifically, for a target image-text pair $(I_{i},T_{i})$, after extracting the feature $f(I_{i})$ and $g(T_{i})$, we separately calculate its distance to the respective modality features within the corresponding memory bank:
\begin{equation}
\begin{aligned}
 \mathcal{D}_{i}^{img} & = [{d(f(I_{i}),f_j)},f_j \in Memory]_{j=1}^{M} \\
 \mathcal{D}_{i}^{text} & = [{d(g(T_{i}),g_j)},g_j \in Memory]_{j=1}^{M}
\label{eq3}
\end{aligned}
\end{equation}
where $d(,.,)$ represents the Euclidean Distance, $\mathcal{D}_{i}^{img}$ and $\mathcal{D}_{i}^{text}$ represent the sets of distances comparing the target image feature $f(I_i)$ and text feature $g(T_i)$ to features stored in the memory bank.
Subsequently, we employ \emph{Spearman's rank correlation coefficient}\cite{spearman1961proof} to evaluate the correlation between these two sets of distances, which is then transformed into a soft correspondence.
Our approach is based on a fundamental assumption about the consistency across different modalities: if a target pair $(I_i, T_i)$ matches, then for two other matched pairs $(I_j, T_j)$ and $(I_m, T_m)$, the shorter distance between $(I_i, I_j)$ compared to $(I_i, I_m)$ should correspond to a shorter distance between $(T_i, T_j)$ than between $(T_i, T_m)$.
Therefore, we can estimate the soft correspondence label $y_i^*$ for $(I_i, T_i)$ by using the correlation between the ranks of $\mathcal{D}_{i}^{img}$ and $\mathcal{D}_{i}^{text}$. 

Specifically, for $\mathcal{D}_{i}^{img}$ and $\mathcal{D}_{i}^{text}$, we compute the rank of each element within them to get $\mathcal{R}_i^{img}$ and $\mathcal{R}_i^{text}$.
The rank of an element $s_i$ in a set can be defined as:
\begin{equation}
\begin{aligned}
R(s_i) = \sum_{j=1}^{n} \mathbb{I}(s_j \leq s_i), s_i \in [s_1, s_2, \dots, s_n]
\label{eq4}
\end{aligned}
\end{equation}

Next, we employ the \emph{Pearson correlation coefficient} to measure the degree of association between $\mathcal{R}_i^{img}$ and $\mathcal{R}_i^{text}$. This \emph{Pearson correlation coefficient} for the rank is also referred to as \emph{Spearman's rank correlation coefficient}, which can be expressed as:
\begin{equation}
\begin{aligned}
corre_{i} = &\frac{\sum (a_j - \bar{a})(b_j - \bar{b})}{\sqrt{\sum (a_j - \bar{a})^2 \sum (b_j - \bar{b})^2}} \\
& a_j \in \mathcal{R}_i^{img}, b_j \in \mathcal{R}_i^{text}, 
\label{eq5}
\end{aligned}
\end{equation}
where $\bar{a}$ and $\bar{b}$ are the mean of $\mathcal{R}_i^{img}$ and $\mathcal{R}_i^{text}$. 
The value of $corre_{i}$, ranging from $[-1,1]$, reflects the correlation of $\mathcal{R}_i^{img}$ and $\mathcal{R}_i^{text}$, where -1 denotes a perfect negative correlation, and 1 denotes a perfect positive correlation.
Finally, we normalize it to the $[0,1]$, ensuring it conforms to the format of a soft label:
\begin{equation}
\begin{aligned}
y^*_i = 
\begin{cases}
0 & \text{if } corre_{i} \leq max(0,\mu), \\
1 & \text{if } corre_{i}  > \gamma, \\
\frac{corre_i - max(0,\mu)}{\gamma - max(0,\mu)} & otherwise
\end{cases}
\label{eq6}
\end{aligned}
\end{equation}
where $\gamma$ is the average of the largest 10\% $corre_{i}$, and  $\mu$ is the average of the smallest 1\% $corre_{i}$.
Eq~\ref{eq6} bases on the assumption that, at a minimum, 10\% of the data pairs are correctly matched and at least 1\% are mismatched.
Additionally, Eq~\ref{eq6} adjusts soft correspondence labels of pairs exhibiting a negative correlation to 0.

\begin{algorithm}[t]
  \footnotesize
  \caption{\small REPAIR } 
  \label{algo}
  \begin{algorithmic}
   \REQUIRE Noisy training set $\Tilde{D} = \{(I_{i},T_{i})\}_{i=1}^N$, $A = \{f^A,g^A,S^A\}$ and $B = \{f^B,g^B,S^B\}$, ${Memory}^A$ and ${Memory}^B$\\
    Warmup $(A, B)$, initiate ${Memory}^A$ and ${Memory}^B$\\
   \textbf{for} each epoch \textbf{do}\\
   \quad Calculate $w_i^k$, $k \in \{A, B\}$ \\
   \quad Divide $\Tilde{D}$ into clean subset and noisy subset according to $w_i^k$, \\
   \quad $\Tilde{D}_B=(D_{c}^{A}, D_{n}^{A})$, $\Tilde{D}_A=(D_{c}^{B}, D_{n}^{B})$: \\
    \qquad  $D_{c}^{A} = \{(I_{i},T_{i}) | w_i^A > p\}$, $D_{n}^{A} = \{(I_{i},T_{i}) | w_i^A <= p\}$ \\
    \qquad $D_{c}^{B} = \{(I_{i},T_{i}) | w_i^B > p\}$, $D_{n}^{B} = \{(I_{i},T_{i}) | w_i^B <= p\}$\\
   \quad \textbf{for} each network $k$ in $(A, B)$  \textbf{do}\\
   \qquad \textbf{for} $iter = 1$ \textbf{to} num\_iters \textbf{do} \\
\qquad \quad Sample a mini-batch $(\mathcal{B}_c^k, \mathcal{B}_n^k)$ from $\Tilde{D}_k$ \\

\qquad \quad Obtain the soft correspondence label for $\mathcal{B}_c^k$ according to rank \\
\qquad \quad correlation\\
\qquad \quad Update $Memory^k$ with features of $\mathcal{B}_c^k$\\
\qquad \quad Select pairs satisfying $(w_i^A<\eta, w_i^B<\eta)$ in $\mathcal{B}_n^k$ to conduct\\
\qquad \quad noisy pair half-replacing\\
\qquad \quad Calculate the $L_{clean}$ and $L_{noisy}$ and train the network $k$ \\
\qquad \textbf{end for} \\
\quad \textbf{end for} \\
   \textbf{end for}
  \end{algorithmic}
\end{algorithm}

Our soft correspondence labels benefit from $Memory$ rather than relying on the outputs of $S$, preventing self-accumulating errors caused by faults in the similarity network.
At the start of each training epoch, for both networks $A$ and $B$, we independently generate the soft correspondence for each pair using their respective memory banks, which is then used to train the other network using Eq~\ref{eq1} and Eq~\ref{eq2} as shown in Figure~\ref{fig:pipeline}(a).
Within a mini-batch, the loss by Eq~\ref{eq1} from $D_c^k$ is denoted as $L_{clean}$.
\subsection{Noisy Pair Half-replacing}
Previous methods adjusted the margin of the triplet loss by estimating the soft correspondence of the pairs. This strategy is effective for matched pairs and weakly matched pairs.
However, for entirely mismatched pairs, even if a small margin is allocated for their low soft correspondence, \textit{i.e.}, $\hat{\alpha}_i = 0$ when $y_i^*$ is $0$ according to Eq~\ref{eq2}, the triplet loss in Eq~\ref{eq1} might still be detrimental.
For instance, given a totally mismatched pair $(I_n, T_n)$, and coincidentally, within the batch, there is an image 
$I_c$ that closely matches the description of $T_n$.
Clearly, existing hard negative mining would easily select $(I_c, T_n)$ as a negative pair. However, when facing a completely mismatched false positive $(I_n, T_n)$, requiring the similarity of $(I_n, T_n)$ to be higher than $(I_c, T_n)$ would harm the model.
Our extensive experiments on the NCR\cite{huang2021learning} method also reveal that integrating $D_{n}^{k}$ into the training phase reduces performance, which is especially pronounced in high-noise environments.
Although method\cite{han2023noisy} proposes abandoning the pairs with low soft correspondence, in our view, this is a suboptimal approach since it doesn't fully utilize the available information.
Thus, based on the memory bank, we propose our noisy pair half-replacing (NPR) to solve the problem.

For a pair $(I_n, T_n)$ nearly certain to be mismatched, our approach is to search the memory bank for the appropriate feature and substitute one of its modalities.
Taking the replacement of the image modality as an example, for the text $T_n$ and corresponding text feature $g(T_{n})$, we aim at finding an image feature $\hat{f}_n$ from the memory bank and building a new feature pair $(\hat{f}_n, g(T_{n}))$ to replace the original mismatch pair.
A suitable $\hat{f}_n$ must satisfy two conditions:1) $\hat{f}_n$ and $g(T_n)$ should exhibit a high degree of soft correspondence, and 2) the corresponding text of $\hat{f}_n$ in memory bank should be similar to $T_n$.  

\begin{figure}[t]
	\includegraphics[width=0.96\linewidth]{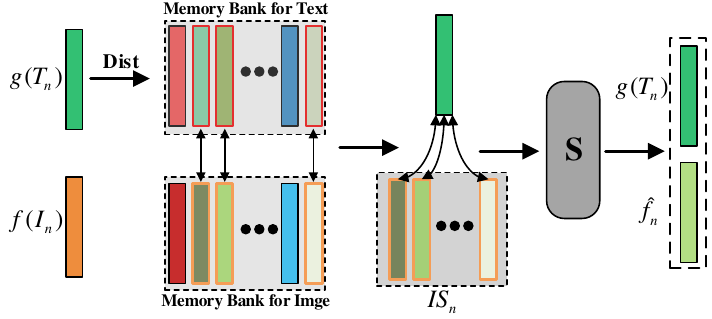}
	\caption{The illustration of the NPR, showing an example to replace the image modality.}  
	\label{fig:replace}
\end{figure}

Thus, we first search the memory bank to find a subset of texts similar to $T_n$:
\begin{equation}
\begin{aligned}
{TS}_n = \{ g_j \ | \ d(g(T_n), g_j) \in \text{topK}(\mathcal{D}_{n}^{text}) \}
\label{eq7}
\end{aligned}
\end{equation}
where $\text{topK}(\mathcal{D}_{n}^{text})$ represents the set of the smallest $K$ distances taken from $\mathcal{D}_{n}^{text}$.
Subset ${TS}_n$ can be considered as features from textual descriptions close to $T_n$.
Hence, based on the almost correctly matched $(f_j, g_j)$ in $Memory$, the images corresponding to ${TS}_n$ might contain visuals that align closely with the $T_n$.
The image feature set ${IS}_n$ corresponding to ${TS}_n$ can be expressed as:
\begin{equation}
\begin{aligned}
{IS}_n = \{ f_j \ | \ g_j \in \{TS_n\} , (f_j, g_j) \in Memory\}
\label{eq8}
\end{aligned}
\end{equation}
Then, we select the image feature with the highest matching degree from them by similarity network $S$:
\begin{equation}
\begin{aligned}
\hat{f}_n = \arg\max_{f_j} S(f_j, T_n), f_j \in IS_n
\label{eq9}
\end{aligned}
\end{equation}

Thus, we can obtain a new feature pair $(\hat{f}_n, g(T_n))$ with a better matching degree to replace the original completely mismatched feature pair.
Similarly, when we replace the text feature $g(T_n)$, we can also obtain $(f(I_n),\hat{g}_n)$.
Since they are feature pairs, we can directly employ Eq~\ref{eq1}, Eq~\ref{eq2}, and Eq~\ref{eq6} to compute the loss, which is denoted as $L_{noisy}$.

In a a batch, we select the pair that satisfy the conditions ($w_i^A < \eta, w_i^B < \eta)$ to conduct NPR, where $\eta$ is the threshold. 
This selection approach can efficiently identify mismatched data pairs with high accuracy, as will be shown in the supplementary material.
We balance $L_{clean}$ and $L_{noisy}$ by setting a weight $\tau$ in a mini-batch:
\begin{equation}
\begin{aligned}
L = L_{clean} + \tau L_{noisy}
\label{eq9}
\end{aligned}
\end{equation}
Finally, we summarizes the pipeline of our proposed REPAIR at Algorithm~\ref{algo}.

\section{Experiments}
In the experiment section, we validate the effectiveness of our method across three image-text matching datasets, \textit{i.e.}, Flickr30K\cite{young2014image}, MS-COCO\cite{lin2014microsoft}, and CC152K\cite{sharma2018conceptual}. Both Flickr30K and MS-COCO are manually annotated datasets, from which we introduce simulated noise by randomly shuffling the captions of images for a specific percentage, denoted as noise rate. And CC152K\cite{sharma2018conceptual, huang2021learning} originates from a real-world noisy environment.

\subsection{Datasets And Implementation Details}
\textbf{Flickr30K}\cite{young2014image} consists of 31,000 images retrieved from the Flickr website, each with five associated annotations.
Following the setting in \cite{lee2018stacked,huang2021learning}, we train with 29,000 images, validate with 1,000 images, and test with another 1,000 images.

\textbf{MS-COCO}\cite{lin2014microsoft} is a collection of 123,287 images.
Within this collection, 113,287 images are split for training, 5,000 for validating, and another 5,000 for testing.
Similar to Flickr30K, every image is accompanied by five textual annotations.

\textbf{CC152K} is a subset segregated from Conceptual Captions\cite{sharma2018conceptual}.
Conceptual Captions, a dataset automatically collected from the internet and comprising 3.3 million image-text pairs, has been shown in \cite{sharma2018conceptual} to contain approximately 3\% to 20\% of samples that are either misaligned or inaccurately described.
From Conceptual Captions, Huang \etal randomly selected 150,000 images for training, 1,000 images for validation, and another 1,000 images for testing to assemble CC152K\cite{huang2021learning}.
We will directly adopt the data selection and partitioning from \cite{huang2021learning} for a fair comparison.

For evaluation, we employ the recall at $K$ ($R$ @ $K$) to measure the performance, which presents the proportion of the queries with the correct item in the closest $K$ retrieved results.
The results of  R@1, R@5, and R@10 are reported in the experiments. 
Following methods\cite{huang2021learning,yang2023bicro,han2023noisy}, the average similarity across networks $A$ and $B$ is the test stage's evaluative criterion.
Across all three datasets, the model exhibiting the best performance on the validation set is selected for testing.

\begin{table*}[ht]
\caption{Performance comparison (R@K(\%)  of image-text retrieval on Flickr30K and MS-COCO 1K. The results of R@1, R@5, R@10 and sum of them are reported. Methods marked by * represent re-implementations using publicly available code.}
\label{tab:perfor}
\begin{center}
\resizebox{\textwidth}{!}{
\begin{tabular}{l|c|ccc|ccc|c|ccc|ccc|c}
\toprule
\multirow{2}{*}{}
 &  & \multicolumn{7}{c|}{Flickr30K} & \multicolumn{7}{c}{MS-COCO 1K}\\
  &  & \multicolumn{3}{c|}{Image to Text} & \multicolumn{3}{c|}{Text to Image} &  &\multicolumn{3}{c|}{Image to Text} & \multicolumn{3}{c|}{Text to Image} &\\
\rule[-1ex]{0pt}{3.5ex}  Noise & Methods & R@1 & R@5 & R@10 & R@1 & R@5 & R@10 & rSum & R@1 & R@5 & R@10 & R@1 & R@5 & R@10 & rSum  \\
\midrule
  \multirow{8}{*}{20\%} 
  & SCAN &56.4 & 81.7 & 89.3 & 34.2 & 65.1 & 75.6 & 402.3 & 28.9 & 64.5 & 79.5 & 20.6 & 55.6 & 73.5 & 322.6\\
  & SGR &61.2 & 84.3 & 91.5 & 44.5 & 72.1 & 80.2 & 433.8 & 49.1 
& 83.8 & 92.7 & 42.5 & 77.7 & 88.2 & 434.0\\
  & VSE$\infty$ & 69.0 & 89.2 & 94.8 & 48.8 & 76.3 & 83.8 & 461.9 & 73.5 & 93.3 & 97.0 & 57.4 & 86.5 & 92.8 & 500.5\\
  & NCR* & 77.5  & 93.3 & 96.1 & 58.3 & 82.9 & 88.8 & 496.9 & 77.7 & 95.8 & 98.3 & 62.5 & 89.6 & 95.3 & 519.2\\
    & DECL & 74.5 & 92.9 & 97.1 & 53.6 & 79.5 & 86.8 & 484.4 & 75.6 & 95.1 & 98.3 & 59.9 & 88.3 & 94.7 & 511.9\\
    & BiCro* & 75.4 & 92.8 & 96.5 & 56.8 & 81.1 & 87.8 & 490.4 & 75.5 & 95.2 & 98.1 & 60.5 & 88.8 & 95.1 & 513.2\\
& MSCN & 77.4 & 94.9 & \textbf{97.6} & \textbf{59.6} & 83.2 & 89.2 & 501.9 & 78.1 & \textbf{97.2} & \textbf{98.8} & \textbf{64.3} & \textbf{90.4} & \textbf{95.8} & \textbf{524.6}\\
& \textbf{REPAIR} & \textbf{79.2} & \textbf{95.0} & 96.9 & 59.4 & \textbf{84.4} & \textbf{89.5} & \textbf{504.4}  & \textbf{78.3} & 96.8 & 98.3 & 62.5 & 89.8 & 95.5& 521.2\\
\midrule
  \multirow{8}{*}{40\%} 
  & SCAN &29.9 & 60.5 & 72.5 & 16.4 & 38.5 & 48.6 & 266.4 & 30.1 & 65.2 & 79.2 & 18.9 & 51.1 & 69.9 & 314.4\\
  & SGR & 47.2 & 76.4 & 83.2 & 34.5 & 60.3 & 70.5 & 372.1 & 43.9 & 78.3 & 89.3 & 37.0 & 72.8 & 85.1 & 406.4\\
  & VSE$\infty$ & 30.2 & 58.3 & 70.2 & 22.3 & 49.6 & 62.7 & 293.3 & 53.3 & 84.3 & 92.1 & 31.4 & 63.8 & 75.0 & 399.9\\
  & NCR* & 74.3 & 92.3 & 95.7 & 55.0 & 81.0 & 87.8 & 486.1 & 76.0 & 95.1 & 98.4 & 60.6 & 88.6 & 94.9 & 513.6\\
    & DECL & 69.0 & 90.2 & 94.8 & 50.7 & 76.3 & 84.1 & 465.1 & 73.6 & 94.6 & 97.9 & 57.8 & 86.9 & 93.9 & 504.7\\
    & BiCro* & 73.5 & 91.8 & 95.6 & 54.3 & 79.1 & 86.5 & 480.8 & 75.5 & 95.2 & 98.0 & 60.4 & 87.9 & 94.2 & 511.2\\
& MSCN & 74.4 & 94.4 & 96.9 & \textbf{57.2} & 81.7 & 87.6 & 492.2 & 74.8 & 94.9 & 98.0 & 60.3 & 88.5 & 94.4 & 510.9\\
& \textbf{REPAIR} & \textbf{75.2} & \textbf{94.6} & \textbf{97.3} & 56.0 & \textbf{81.8} & \textbf{87.9} & \textbf{492.8}  & \textbf{76.4} & \textbf{95.4} & \textbf{98.4} & \textbf{61.0} & \textbf{88.8} & \textbf{95.0} & \textbf{515.0}\\
\midrule
  \multirow{8}{*}{60\%} 
  & SCAN &16.9 & 39.3 & 53.9 & 2.8 & 7.4 & 11.4 & 131.7 & 27.8 & 59.8 & 74.8 & 16.8 & 47.8 & 66.4 & 293.4\\
  & SGR &28.7 & 58.0 & 71.0 & 23.8 & 49.5 & 60.7 & 291.7 & 37.6 &73.3 &86.3 & 33.8 & 68.6 & 81.7 & 381.3\\
  & VSE$\infty$ & 18.0 & 44.0 & 55.7 & 15.1 & 38.5 & 51.8 & 223.1 & 33.4 & 64.8 & 79.1 & 26.0 & 60.1 & 76.3 & 339.7\\
  & NCR* & 70.1 & 90.0 & 95.0 & 51.1 & 77.1 & 84.4 & 467.7 & 72.8 & 94.0 & 97.4 & 57.8 & 87.1 & 94.0 & 503.1\\
    & DECL & 64.5 & 85.8 & 92.6 & 44.0 & 71.6 & 80.6 & 439.1 & 69.7 & 93.4 & 97.5 & 54.5 & 85.2 & 92.6 & 492.9\\
    & BiCro* & 69.6 & 90.9 & 94.1 & 50.5 & 76.4 & 83.7 & 465.2 & \textbf{74.7} & 94.2 & \textbf{97.9}  & 58.8 & 86.9 & 93.7 & 506.2\\
& MSCN & 70.4 & 91.0 & 94.9 & \textbf{53.4} & 77.8 & 84.1 & 471.6 & 74.4 & \textbf{95.1} & \textbf{97.9} & \textbf{59.2} & 87.1 & 92.8  & \textbf{506.5}\\
& \textbf{REPAIR} & \textbf{72.3} & \textbf{91.4} & \textbf{95.4} & 52.6 & \textbf{79.1} & \textbf{86.1} & \textbf{476.9}  & 73.7 & 94.4 & \textbf{97.9} & 58.6 & \textbf{87.3} & \textbf{94.1} & 506.0\\
\bottomrule
\end{tabular}
}
\end{center}
\end{table*}

As a general framework, REPAIR can be directly integrated into existing cross-modal methods to improve performance without additional data preparation or adaptation of network structures.
Following \cite{huang2021learning}, we select SGR\cite{diao2021similarity} as our backbone, which includes projection module and similarity network.
As for $f$ and $g$, we adopt the setting in \cite{huang2021learning,yang2023bicro,han2023noisy} for a fair comparison, \textit{i.e.}, a fully connected layer for $f$ and a Bi-GRU\cite{schuster1997bidirectional} for $g$. 
For all three datasets, we set the batchsize as 128.
We train the model by the Adam optimizer\cite{kingma2014adam} with a 0.0002 initial learning rate.
As for Flickr30K and MS-COCO, the training stage lasts 40 epochs, during which the learning rate decays by 0.1 after 30 epochs.
As for CC152K, we set the training stage as 20 epochs and decay the learning rate by 0.1 after 10 epochs. 
We set the memory bank size $M$ as 4096 and $\eta$ as 0.25. 
For the remaining hyperparameters, $\alpha$ is set as 0.2, $p$ as 0.5, $m$ at 10, $\tau$ at 0.15, and $K$ at 32.
All experiments are conducted using PyTorch 2.0 and on a single RTX 4090 GPU.

\subsection{Comparison With The Other Methods}
In this section, we compare our method with the other cross-modal methods, including: SCAN\cite{lee2018stacked}, SGR\cite{diao2021similarity}, VSE$\infty$\cite{chen2021learning}, NCR\cite{huang2021learning}, DECL\cite{qin2022deep}, BiCro\cite{yang2023bicro}, MSCN\cite{han2023noisy}.
Among these, the latter four methods are specially designed to address the issue of noisy correspondence.
For a fair comparison, we report the performance of NCR, DECL, BiCro, and MSCN based on SGR-backbone, eliminating performance biases due to variations in network architecture.
As for Flickr30K and MS-COCO, we introduce synthetic noise by randomly shuffling the training images and captions for a certain noise rate.
Specifically, 20\%, 40\%, and 60\% noise rates are reported in our experiment.

\begin{table}[ht]
\caption{Performance comparison (R@K(\%)  of image-text retrieval on CC152K. The results of R@1, R@5, R@10 and sum of them are reported. Methods marked by * represent re-implementations using publicly available code.}
\label{tab:performance_cc152}
\begin{center}
\resizebox{\columnwidth}{!}{
\begin{tabular}{c|ccc|ccc|c}
\toprule
  & \multicolumn{3}{c|}{Image to Text} & \multicolumn{3}{c|}{Text to Image} & \\
\rule[-1ex]{0pt}{3.5ex}  Methods & R@1 & R@5 & R@10 & R@1 & R@5 & R@10 & rSum \\
\midrule
  SCAN &30.5 & 55.3 & 65.3 & 26.9 & 53.0 & 64.7 & 295.7 \\
  SGR &11.3 & 29.7 & 39.6 & 13.1 & 30.1 & 41.6 & 165.4 \\
  VSE$\infty$ &34.0 & 64.5 & 77.0 & 12.9 & 19.2 & 21.6 & 229.2 \\
  NCR* & 39.1 & 65.4 & 75.2 & 40.6 & 66.4 & 74.5 & 361.2 \\
  DECL & 36.2 & 63.6 & 73.2 & 37.1 & 63.6 & 73.7 & 347.4 \\
  BiCro* & 37.6 & 61.4 & 69.8 & 38.5 & 63.2 & 72.3 & 342.8 \\
  MSCN & 40.1 & 65.7 & \textbf{76.6} & \textbf{40.6} & 67.4 & 76.3 & 366.7 \\
  REPAIR & \textbf{40.5} & \textbf{67.7} & 76.1 & 40.3 & \textbf{68.2} & \textbf{76.4} & \textbf{369.2} \\   
\bottomrule
\end{tabular}
}
\end{center}
\end{table}

\begin{table*}[ht]
\caption{Ablation studies on Flickr30K with 40\% noise.}
\label{tab:ablation_study}
\begin{center}
\begin{tabular}{cccc|ccc|ccc|c}
\toprule
 \multicolumn{4}{c|}{method} & \multicolumn{3}{c|}{Image to Text} & \multicolumn{3}{c|}{Text to Image} \\
\rule[-1ex]{0pt}{3.5ex}  hard & RC  & drop & NPR & R@1 & R@5 & R@10 & R@1 & R@5 & R@10 &rSum\\
\midrule
 \checkmark&  &  &  &72.9 & 92.5 & 95.7 & 54.7 &79.7 & 86.8 & 482.3  \\
  \checkmark&  & \checkmark &  & 74.3 & 92.6 & 96.0  & 55.9 & 80.2 & 87.3 & 486.3\\
  & \checkmark & \checkmark &  & 75.0 & 93.4 & 97.1 & 55.7 & \textbf{82.1} & \textbf{88.3} & 491.6\\
& \checkmark &  & \checkmark &\textbf{75.2} & \textbf{94.6} & \textbf{97.3} & \textbf{56.0} &81.8 & 87.9 & \textbf{492.8}\\

\bottomrule
\end{tabular}
\end{center}
\end{table*}

\begin{table}[ht]
\caption{Performance comparison (R@K(\%)  of image-text retrieval on MS-COCO 5K. The results of R@1, R@5, R@10 and sum of them are reported. Methods marked by * represent re-implementations using publicly available code.}
\label{tab:performance_ms5k}
\begin{center}
\resizebox{\columnwidth}{!}{
\begin{tabular}{c|c|ccc|ccc|c}
\toprule
   & & \multicolumn{3}{c|}{Image to Text} & \multicolumn{3}{c|}{Text to Image} & \\
\rule[-1ex]{0pt}{3.5ex} Noise & Methods & R@1 & R@5 & R@10 & R@1 & R@5 & R@10 & rSum \\
\midrule
\multirow{4}{*}{20\%}
  & NCR* & 56.3 & 83.4 & 90.8 & 40.8 & 69.7 & 80.4 & 421.4 \\
  & BiCro* & 53.3 & 81.8 & 89.7 & 38.9 & 68.1 & 79.1 & 410.9 \\
  & MSCN & 57.1 & 84.0 & 90.8 & \textbf{41.0} & 69.8 & 80.3 & 423.0 \\
  & \textbf{REPAIR} & \textbf{57.6} & \textbf{84.1} & \textbf{91.4} & \textbf{41.0} &\textbf{70.2} & \textbf{80.7} & \textbf{425.0} \\
  \midrule
  \multirow{4}{*}{40\%}
  & NCR* &55.1 & 82.0 & 90.0 & 39.1 & 68.2 & 78.7 & 413.1 \\
  & BiCro* & 50.7 &79.4 & 88.2 & 37.1 & 66.2 & 77.3 & 398.9 \\
  & MSCN & - & - & - & - & - & - & - \\
  & \textbf{REPAIR} & \textbf{55.5} & \textbf{82.8} & \textbf{90.1} & \textbf{39.2} & \textbf{68.6} & \textbf{79.2} & \textbf{415.4} \\
  \midrule
    \multirow{4}{*}{60\%}
    & NCR* & 50.4 & 78.5 & 87.9 & 36.3 & 65.2 & 76.6 & 394.9 \\
  & BiCro* & 47.5 & 76.8 & 85.7 & 34.2 & 62.9 & 74.5 & 381.6 \\
  & MSCN & - & - & - & - & - & - & - \\
    &\textbf{REPAIR} & \textbf{52.5} & \textbf{79.9} & \textbf{88.2} & \textbf{36.8} & \textbf{65.9} & \textbf{77.0} & \textbf{400.3} \\
\bottomrule
\end{tabular}
}
\end{center}
\end{table}

As shown in Table~\ref{tab:perfor}, we compare our method with other noisy correspondence methods on Flickr30K and MS-COCO.
The results of R@1, R@5, R@10, and the sum of these ranks (denoted as rSum) are reported for exhaustive comparison.
For MS-COCO, we compare the results by averaging over 5 folds of 1K test images (denotes as MS-COCO 1K) following \cite{huang2021learning, yang2023bicro, han2023noisy}.
For Flickr30K dataset, our method exhibits the best performance across all noise settings.
Specifically, under 20\%, 40\%, and 60\% noise rates, the rSum of REPAIR outperform the previous best-performing method MSCN by 2.5\%, 0.6\%, and 5.2\%, respectively.
It must be noted that MSCN\cite{han2023noisy} requires an additional clean dataset for training the meta network, which undoubtedly gives MSCN a distinct advantage.
As for MS-COCO 1K, our method also achieves competitive performance.
We also report the results of MS-COCO dataset on 5K test images (denoted as MS-COCO 5K), which is shown on Table~\ref{tab:performance_ms5k}.
Compared with other noisy correspondence methods, our REPAIR achieve the best performance.
The results for the real-world dataset CC152K are as reported in Table~\ref{tab:performance_cc152}.
Compared with other noisy correspondence methods NCR, BiCro, DECL and MSCN, our REPAIR achieved advantages of 8.0\%, 26.4\%, 21.8\%, and 2.5\% in rSum, respectively.
This indicates that our REPAIR method still possesses strong competitiveness and achieves good performance in real-world noisy environments.

To demonstrate generalization capability of REPAIR across different backbones, we also select VSE$\infty$\cite{chen2021learning} as the backbone and re-implement our method and the baseline NCR on it.
In contrast to backbone SGR\cite{diao2021similarity}, VSE$\infty$ has fewer parameters and lacks a similarity network, thereby increasing the challenge of achieving high performance.
The rSum of three datasets are reported in Table~\ref{tab:performance_vse}.
Our approach, utilizing the VSE$\infty$ backbone, outperformed NCR in all three datasets, as can be observed.

\begin{table}[ht]
\caption{Performance of NCR and REPAIR based on VSE$\infty$ backbone. The rSums (R@1, R@5, R@10) of Flickr30K, MS-COCO 1K and CC152K are reported.}
\label{tab:performance_vse}
\begin{center}
\resizebox{\columnwidth}{!}{
\begin{tabular}{c|ccc|ccc|c}
\toprule
  methods& \multicolumn{3}{c|}{Flickr30K} & \multicolumn{3}{c|}{MS-COCO 1K} & CC152K\\
\rule[-1ex]{0pt}{3.5ex}  noise & 0.2 & 0.4 & 0.6 & 0.2 & 0.4 & 0.6 &  \\
\midrule
  NCR-VSE$\infty$ & 445.9 & 400.8 & 332.2 & 502.7 & 494.5 & 475.8 & 345.5\\
  REPAIR-VSE$\infty$ &\textbf{452.3} & \textbf{405.3} & \textbf{338.9} & \textbf{508.1} & \textbf{497.3} & \textbf{480.4} & \textbf{348.9} \\  
\bottomrule
\end{tabular}
}
\end{center}
\vspace{-0.5cm}
\end{table}

\begin{figure}[t]
	\includegraphics[width=0.96\linewidth]{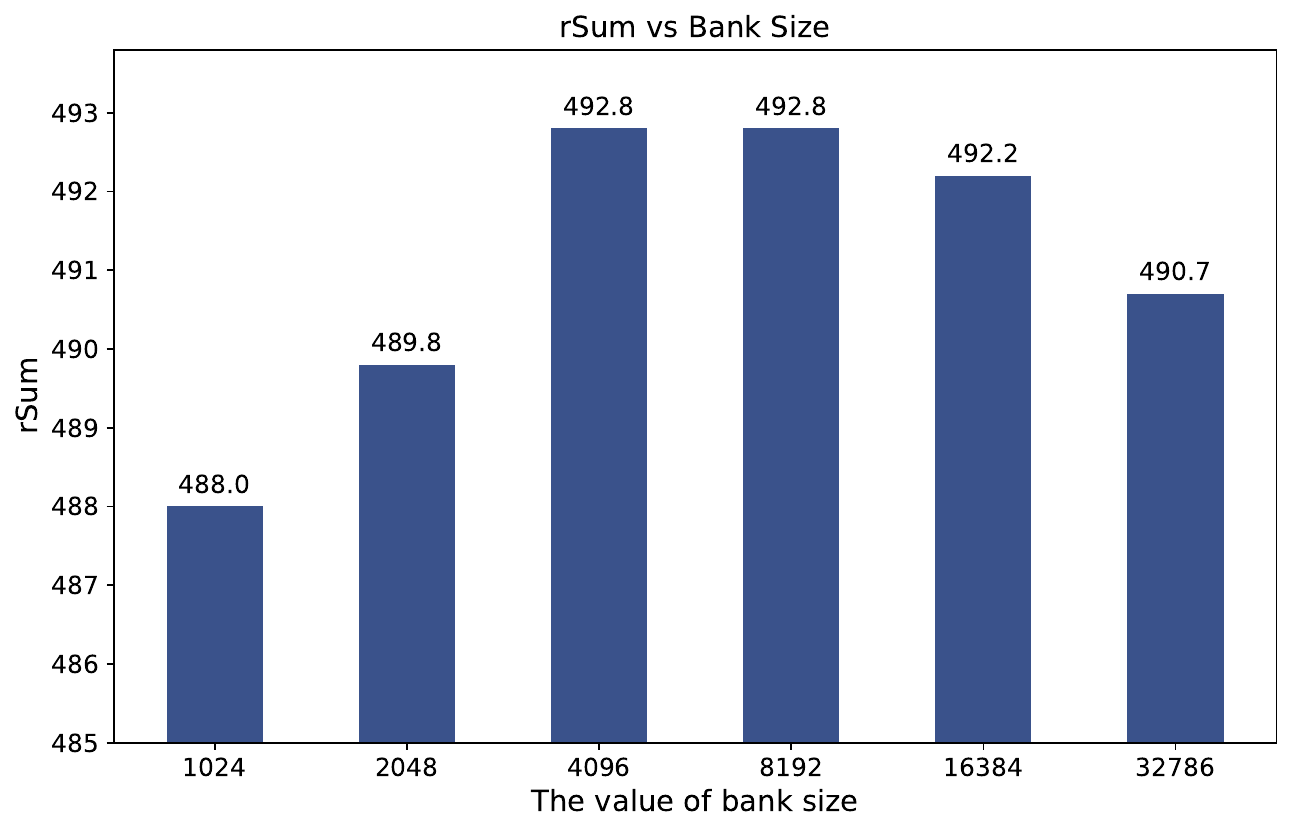}
	\caption{The performance on Flickr30K with varying bank size. The experiments are conducted under 40\% noise rate.}  
	\label{fig:parameter_analyse}
\end{figure}

\subsection{Ablation Study}
As shown in Table~\ref{tab:ablation_study}, we conduct experiments to validate the effectiveness of each component in our REPAIR framework.
We design two additional modules for the purpose of comparison:
\emph{hard} implies that the labels for $D_{c}^{k}$ and $D_{n}^{k}$ are assigned as 1 and 0, respectively. 
This results in a uniform margin $\alpha$ for all data pairs originating from $D_{c}^{k}$, with the margin for data pairs from B being $D_{n}^{k}$.
\emph{drop} means discarding data pairs in $D_{n}^{k}$ directly and only adopt pairs in $D_{c}^{k}$ in the current training epoch.
Comparing the first and second rows, this once again confirms that discarding the mismatched data pairs is indeed an effective strategy.
This is because mismatched data pairs typically generate larger losses and mislead the neural network.
Comparison between the second and third rows, proving the effectiveness of RC module.
And the comparison between the third and fourth rows demonstrates that NPR can yield greater improvements than \emph{drop}.

\subsection{Memory Usage And Training Time}
As shown in Table~\ref{tab:memory_and_time}, we report the training time and extra memory usage of our method on RTX 4090.
The key factors affecting training time and memory usage are the choice of backbone and the size of the memory bank.
Hence, we have listed metrics for two backbones, SGR and VSE$\infty$, under various bank size.
The text encoder of SGR is more complex, capable of extracting finer text features from the text, thus requiring more memory to store these features than VSE$\infty$.
Additionally, from our observation, the majority of the memory in SGR's memory bank is allocated for storing text features. This is because image features are relatively simple in comparison, and the text features in SGR are related to the length of sentences.
As for the time required, REPAIR takes 28.1 hours and 7.5 hours to train on the SGR and VSR$\infty$ backbones, respectively, at a bank size of 4096.
Further increasing the bank size will lead to longer training times due to the additional calculations and comparisons of distances between features.
Therefore, the bank size should be set to an appropriate value to avoid excessive computational costs.

\begin{table}[t]
\caption{The training time and extra memory usage for the memory bank of different bank size and backbone on Flickr30K. Experiments are conducted on a single RTX 4090. MB and GB correspond to the memory units Megabyte and Gigabyte, respectively, while 'h' is the abbreviation for the time unit hour.}
\label{tab:memory_and_time}
\begin{center}
\resizebox{\columnwidth}{!}{
\begin{tabular}{cccc}
\toprule
\rule[-1ex]{0pt}{3.5ex} backbone & bank size & extra memory usage&training time \\
\midrule
SGR & 4096 & 1.85GB & 28.1h \\
SGR & 8192 & 3.63GB & 36.8h \\
VSE$\infty$ & 4096 & 81MB & 7.5h \\
VSE$\infty$ & 8192 & 137MB & 11.5h \\
\bottomrule
\end{tabular}
}
\end{center}
\vspace{-0.8cm}
\end{table}

\subsection{Hyperparameter Analysis}
To verify the effectiveness of each key parameter, we set up comparative experiments on the Flickr30K dataset with a noise rate of 0.4.
The effect of different bank size $M$ is shown in Figure~\ref{fig:parameter_analyse}.
It can be observed that the best results are obtained when $M$ reaches 4096 or 8192, and further increasing the bank size does not yield significant improvements.
Additionally, a greater value of $M$ leads to prolonged training times, as it requires increased comparisons of feature distances between the target pair and features within the memory bank.
There, we set $M$ to 4096 for all experiments by default to balance performance with training time.

\begin{figure}[t]
	\includegraphics[width=0.96\linewidth]{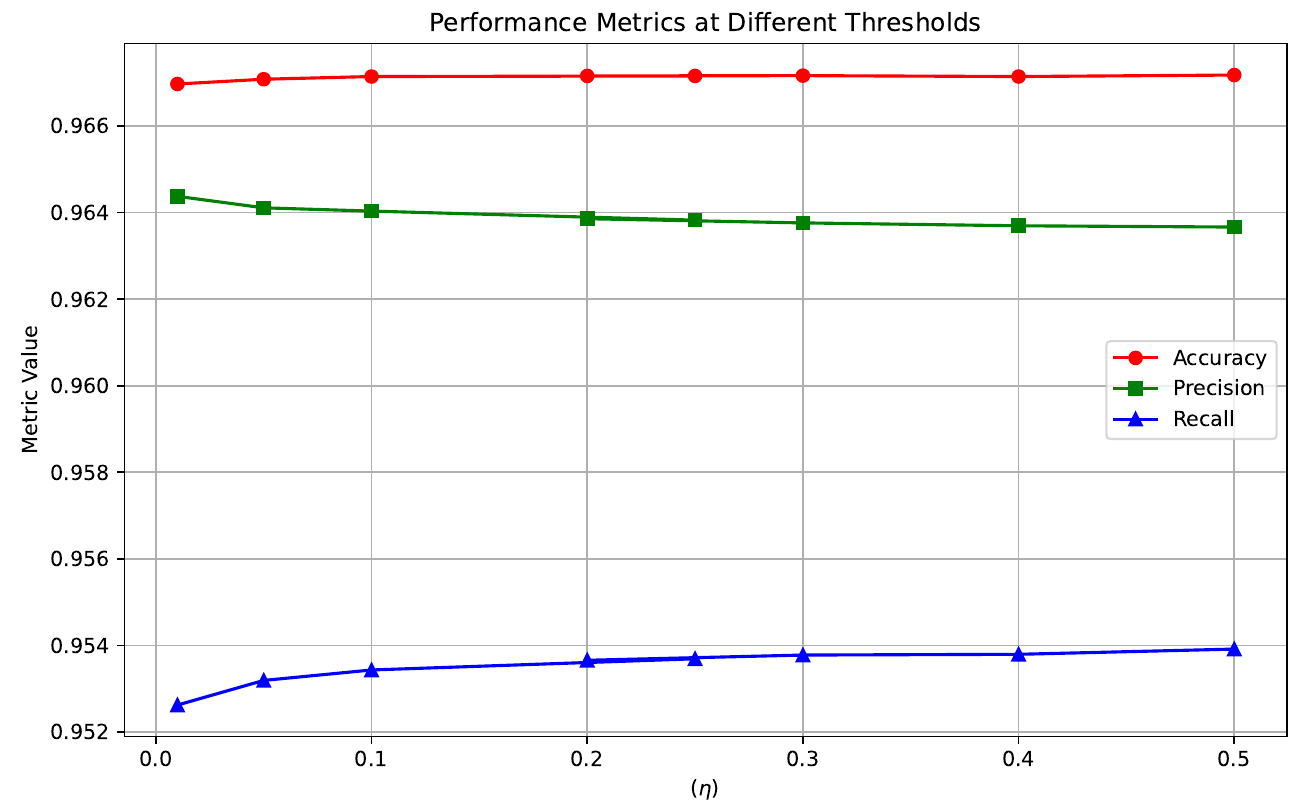}
	\caption{The accuracy, precision and recall on Flickr30K with different value of $\eta$. The experiments are conducted under 40\% noise rate.}  
	\label{fig:parameter_eta}
\end{figure}

The validation of the hyperparameter $\eta$, as shown in Figure~\ref{fig:parameter_eta}, involves setting various $\eta$ values to compare the discrimination capability for noise data in NPR.
The accuracy, precision, and recall rates of identifying mismatched data pairs within the [0,0.5] range for $\eta$ have been documented.
It is observable that $\eta$ demonstrates robust performance across various values, achieving good performance on three metrics at different settings.
This is primarily because we employed the posterior probability of GMM as the discriminative criterion, where GMM exhibits a high skew towards the clean set\cite{han2023noisy,yang2023bicro}.

\begin{figure}[t]
	\includegraphics[width=0.96\linewidth]{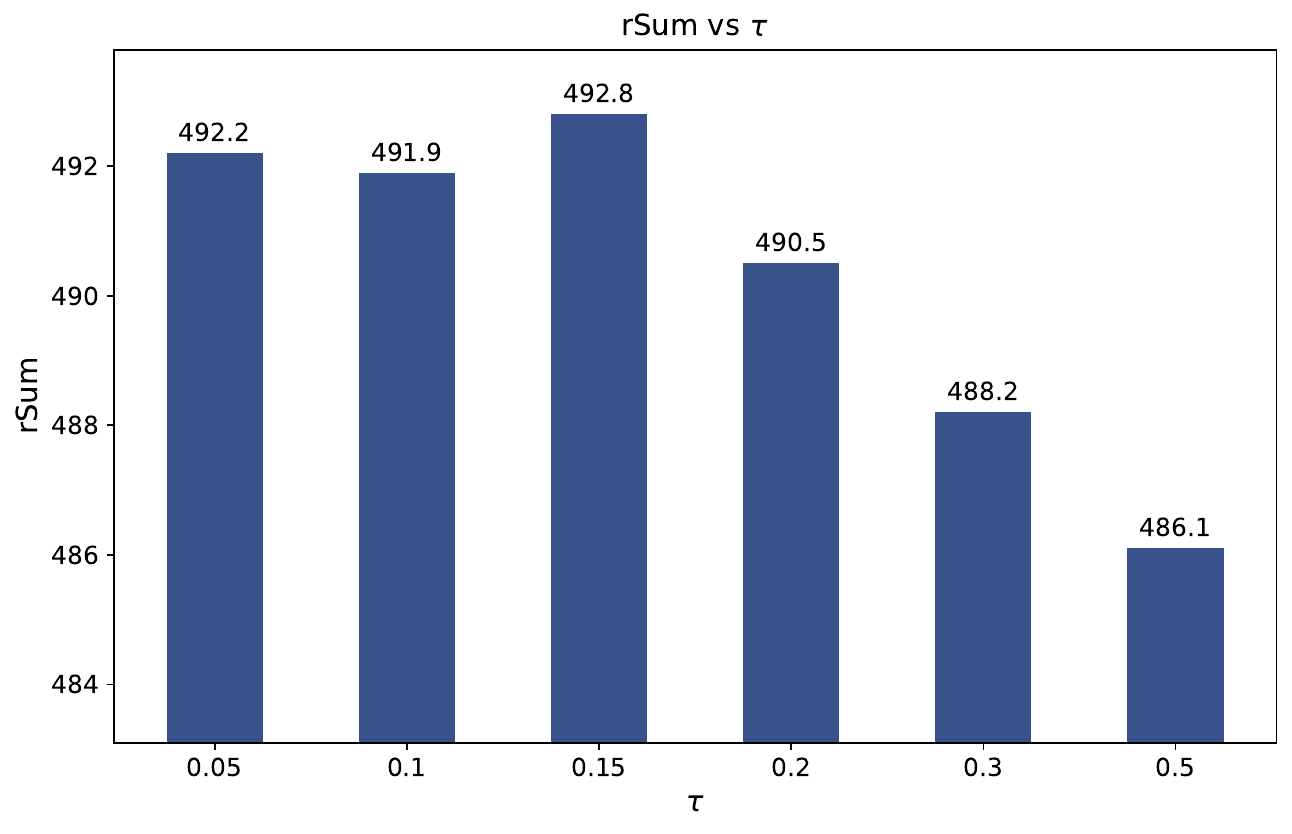}
	\caption{The performance on Flickr30K with different value of $\tau$. The experiments are conducted under 40\% noise rate.}  
	\label{fig:parameter_tau}
\end{figure}

We report the performance of different value of $\tau$ in Figure~\ref{fig:parameter_tau}.
It is evident that values of $\tau$ that are too small or too large result in decreased performance, with the best performance being achieved at a value of 0.15.
When the value of $\tau$ is relatively low, the loss weight for $L_{noisy}$ is reduced, making NPR akin to directly discarding the noisy data.
However, when the weights are higher, although our NPR employs a new feature pair to replace mismatched pair, performance still diminishes under heavy weights.
This is because the newly generated feature set still consists of pseudo pairs, which can be considered an approximation of matching features. When assigned significant weight, it may hinder the neural network's learning of truly matching pairs.

\subsection{Visualization}
\begin{figure}[t]
	\includegraphics[width=0.96\linewidth]{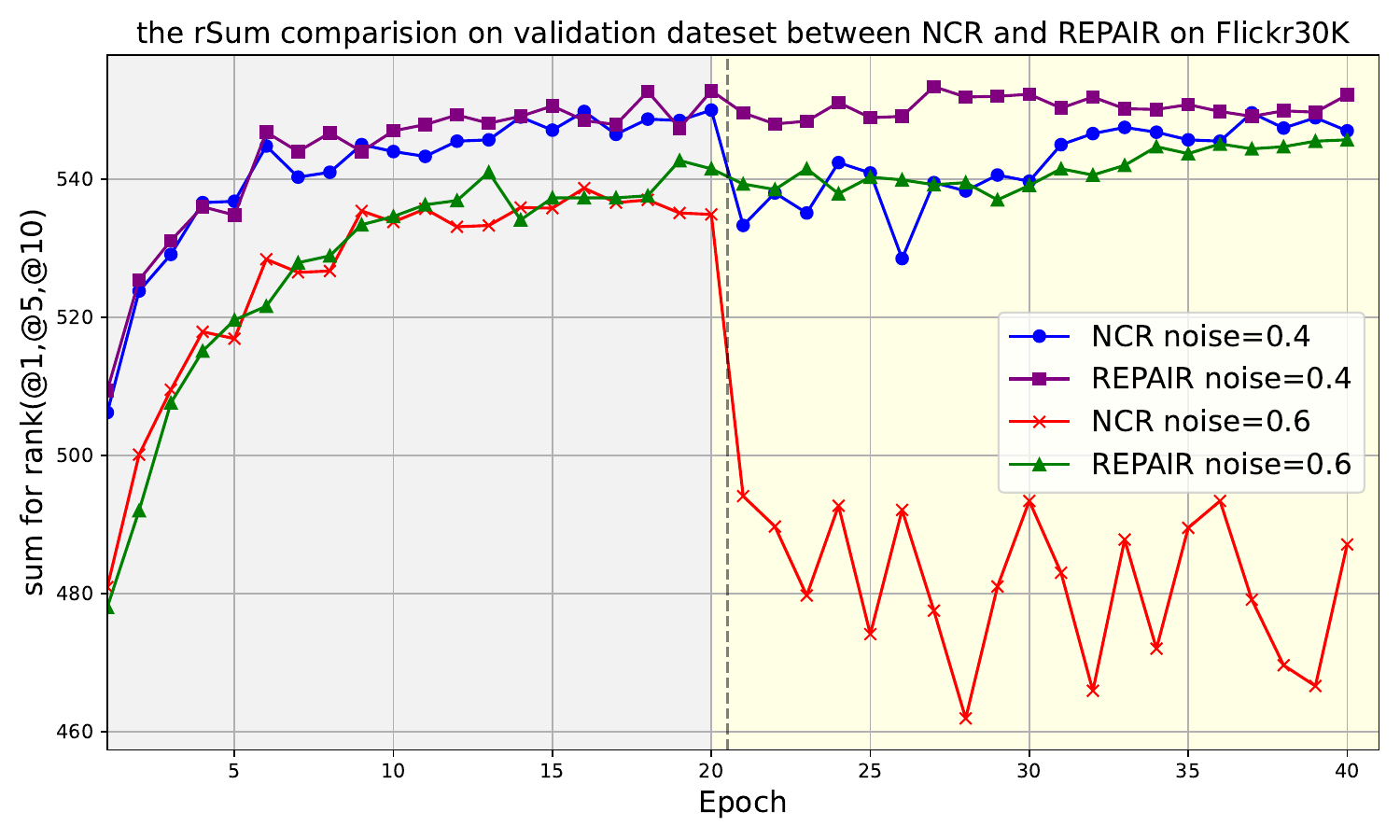}
	\caption{The performance on the validation set of Flickr30K during the training stage. After 20 epochs, the loss from the noise set is incorporated into the training.}  
	\label{fig:ncr_repair_val}
\end{figure}

\begin{figure}[t]
	\includegraphics[width=0.96\linewidth]{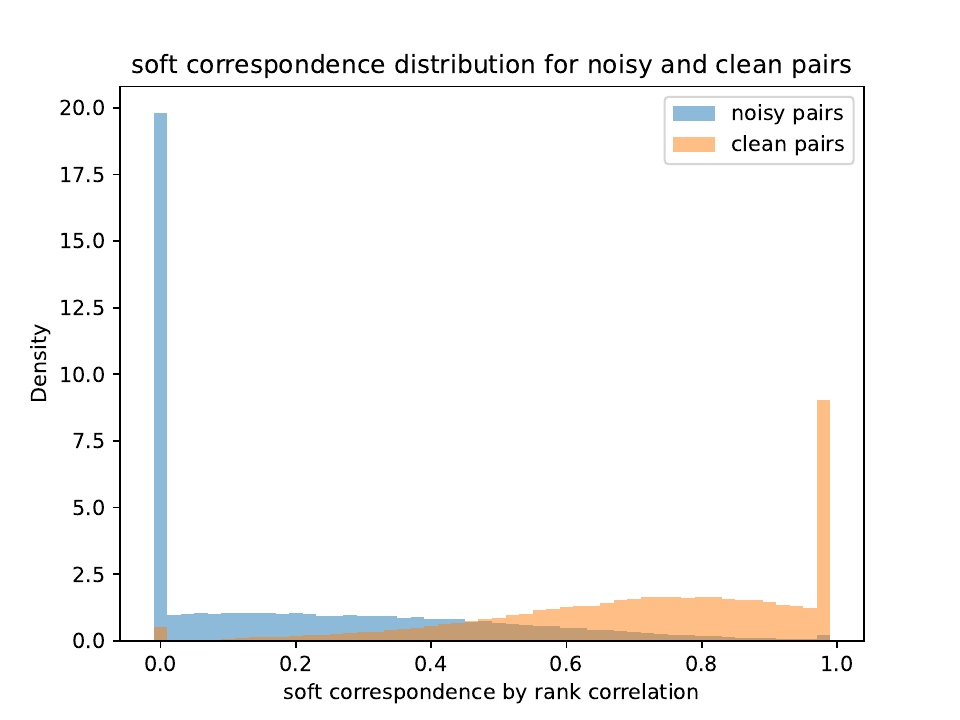}
	\caption{The distribution of soft correspondence $y^*$ predicted by rank correlation on Flickr30K with 40\% noise rate. Noisy pairs and clean pairs are marked in blue and orange, respectively.}  
	\label{fig:density}
\end{figure}

We have visualized the performance variations of REPAIR and NCR on the validation set during the training phase, as shown in Figure~\ref{fig:ncr_repair_val}.
For the first 20 epochs of the model, only the loss from the clean set is utilized, with the loss from the noise set being introduced after 20 epochs.
Comparing the first 20 epochs, it is evident that the RC module achieves better performance than NCR.
After the introduction of the loss from the noisy set, NCR experience a significant decrease in performance, whereas REPAIR did not. Instead, REPAIR enhance its performance further with NPR.
We have also visualized the situation of soft correspondence predicted by Rank Correlation for noisy pairs and clean pairs, as shown in Figure~\ref{fig:density}.
It can be observed that Rank Correlation effectively distinguishes between noisy and clean pairs, which will assist in the dynamic margin adjustment during training.

\begin{figure*}[t]
	\includegraphics[width=0.96\linewidth]{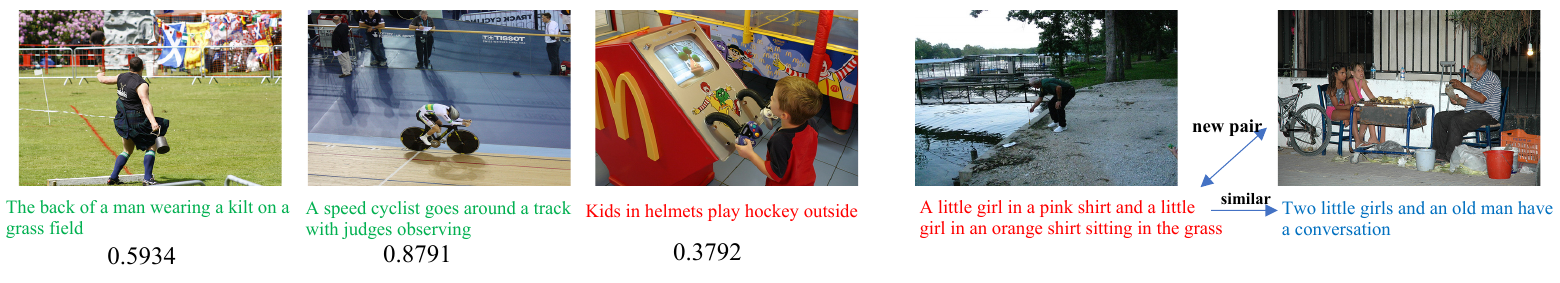}
	\caption{Some examples from Flickr30K dataset by adopting our REPAIR. The text in matched image-text pairs is indicated by green font, whereas the text in completely mismatched pairs is represented by red.} 
	\label{fig:fig_example}
\end{figure*}

We show some visualization examples from Flickr30K by adopting our REPIAR in Figure~\ref{fig:fig_example}.
The first two columns are relatively matching pairs, and REPAIR gives a higher soft correspondence.
The third column is the mismatched pair that has similar semantic relations( kids; play), and REPAIR assigns a moderately low matching degree.
Columns four and five illustrate an application example of NPR.
For completely mismatched data pairs, we find text similar to the original text from the memory bank, then form new data pairs using images corresponding to these similar texts alongside the original text.

\subsection{Discussion}
In this subsection, we discuss the comparison between REPAIR and other noisy correspondence methods, to clearly define the novelty, advantages, and disadvantages of REPAIR.
The key difference between our approach and other noisy correspondence methods lies in employing a memory bank to address noise issues in multimodal field.
Our use of memory bank is not just about integrating a common tool to improve performance in the noisy correspondence area.
Instead, our improvements based on the memory mank specifically address two persistent issues: a). the self-accumulation errors, and b). the inefficient use of noisy data.
In response, we offer Rank Correlation and Noisy Pair Half-replacing as targeted solutions.

Rank correlation utilizes a memory bank for evaluating the matching degree of multimodal data, which is based on the consistency of multimodal sample distance ranks, differing from existing noisy correspondence methods.
In contrast to \textbf{NCR}, rank correlation relies on features instead of results predicted by the similarity network, reducing the risk of cumulative errors in cases of incorrect network predictions.
Compared to another feature relationship based method for noisy correspondence, \textbf{BiCro}, which evaluates the similarity of target sample pairs by the ratio of distances to the nearest sample pair within a batch, our method benefits from obtaining a memory bank. BiCro's limitation arises from potential biases in single-point estimates, particularly when the closest sample pair in the batch significantly differs from the target pair. In contrast, our rank correlation method circumvents this issue by considering the ranking order relationships among a set of samples.
As for Noisy Pair Half-replacing (NPR), this is the first method in noisy correspondence to focus on how to efficiently utilize mismatched data pairs. Instead of simply discarding them, utilizing NPR in a clean memory bank to find highly matching samples can further enhance performance.
Thus, we believe REPAIR makes a significant contribution to the noisy correspondence community.

Regarding the drawbacks, the primary disadvantages of REPAIR are the additional memory usage and inference time, especially when the feature dimensions of the backbone are large.
For instance, the text features in SGR are quite complex, resulting in features of size $[sentence\_length, 36, embedding\_size]$, where $sentence\_length$ and $embedding\_size$ are the length of the target sentence and embedding size of feature, which require larger storage space.
However, this issue does not affect the usability of REPAIR, as our experiments have demonstrated that a very large bank size is not necessary to achieve good performance.
Therefore, this configuration does not result in excessive memory usage, even on very large datasets.
Likewise, due to the avoidance of excessive bank sizes, the extra time required is also within a tolerable limit.

In future work, we will continue to focus on the issue of noisy correspondence, investigating how to efficiently address noise in multimodal contexts.
Currently, most methods evaluate the clean/noisy probability of each sample through loss. However, this method is not capable of handling complex scenarios, such as hard samples being mistaken for noisy ones due to high loss.
Unfortunately, in real-world multimodal matching, the occurrence of hard examples is not uncommon, due to the diversity within modal distributions.
Therefore, in our future work, we will consider how to differentiate between such hard examples and noisy samples, where utilizing a memory bank may offer a viable solution.

\section{Conclusion}
In this paper, we propose a memory-based approach to address the issue of noisy correspondence.
Specifically, we maintain a feature memory bank for the set of clean data pairs.
Based on the idea that \emph{similar relationships in one modality should mirror those in its connected modality}, we evaluate the soft correspondence of target pairs according to \emph{Spearman's rank correlation coefficient}, which measures the correlation between the target sample pairs and the memory bank features across both modalities.
Such soft correspondence is converted into a soft margin, applying stricter constraints to pairs with higher matching degree.
Furthermore, for data pairs with low soft correspondence, we propose a \emph{noisy pair half-replacing} strategy based on memory bank to substitute the feature of one modality, thereby acquiring data with a higher matching degree.
We carry out experiments on both synthetic noise and real-world noisy scenarios, validating the effectiveness of our method.
\bibliography{total.bib}{}
\bibliographystyle{IEEEtran}

%

\vspace{11pt}
\begin{IEEEbiography}[{\includegraphics[width=1in,height=1.25in,clip,keepaspectratio]{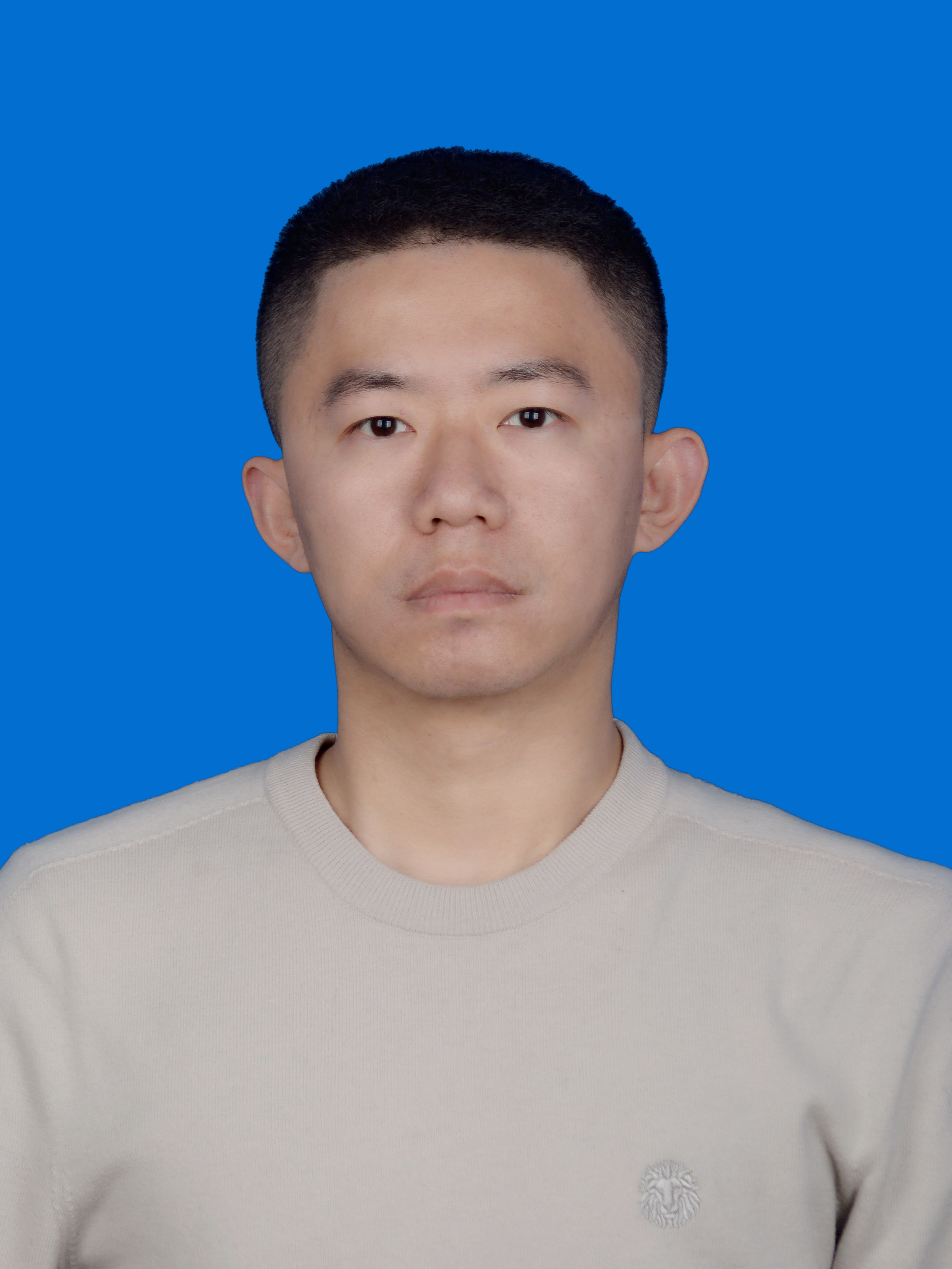}}]{Ruochen Zheng}
received a B.Eng. degree and M.S. degree in automation the School of Artificial Intelligence and
Automation from Huazhong University of Science and Technology, where he is currently pursuing an Ph.D. degree in the School of Artificial Intelligence and Automation.
His current research interests are noisy correspondence, multi-modal matching.
\end{IEEEbiography}

\vspace{11pt}
\begin{IEEEbiography}[{\includegraphics[width=1in,height=1.25in,clip,keepaspectratio]{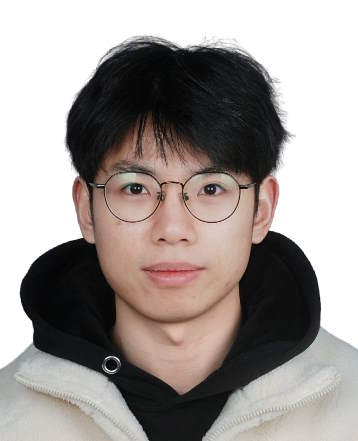}}]{Jiahao Hong}
received a B.Eng. degree from the School of Artificial Intelligence and Automation, Huazhong University of Science and Technology, where he is currently pursuing an M.S. degree with the School of Artificial Intelligence and Automation. His research interests focus on unsupervised person re-identification.
\end{IEEEbiography}

\vspace{11pt}
\begin{IEEEbiography}[{\includegraphics[width=1in,height=1.25in,clip,keepaspectratio]{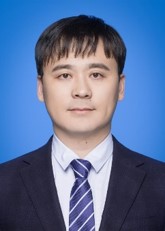}}]{Changxin Gao}
received a Ph.D. degree in pattern recognition and intelligent systems from Huazhong University of Science and Technology in 2010. He is currently a professor at
the School of Artificial Intelligence and Automation, Huazhong University of Science and Technology, China. His research interests are pattern recognition and video analysis.
\end{IEEEbiography}

\vspace{11pt}
\begin{IEEEbiography}[{\includegraphics[width=1in,height=1.25in,clip,keepaspectratio]{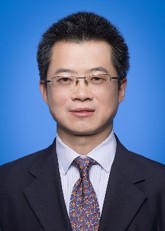}}]{Nong Sang}
received a Ph.D. degree in pattern recognition and intelligent systems from Huazhong University of Science and Technology in 2000. He is currently a professor at the
School of Artificial Intelligence and Automation,
Huazhong University of Science and Technology, China. His research interests include object detection and recognition, object tracking,
image/video semantic segmentation, intelligent
processing, and video analysis.
\end{IEEEbiography}

\vfill




\end{document}